\documentclass{article}
  \pdfoutput=1

\usepackage{arxiv}
\usepackage[utf8]{inputenc} 
\usepackage[T1]{fontenc}    
\usepackage{hyperref}       
\usepackage{url}            
\usepackage{booktabs}       
\usepackage{amsfonts}       
\usepackage{nicefrac}       
\usepackage{microtype}      
\usepackage{lipsum}
\usepackage{graphicx}
\usepackage{amsmath}
\usepackage{microtype}
\usepackage{empheq}
\usepackage{pifont}
\usepackage{amssymb}
\usepackage{mathtools}
\usepackage{adjustbox}
\usepackage{times}
\usepackage{latexsym}

\title{Word2rate: Training and evaluating multiple word embeddings as statistical transitions}

\author{
  Dario Poletti \\
  Engineering Product Development\\
  Singapore University of Technology and Design\\
  \texttt{dario\_poletti@sutd.edu.sg} \\
     \And
  Lin Shaowei \\
  Engineering Systems and Design\\
  Singapore University of Technology and Design\\
  \texttt{shaowei@gmail.com} \\
  \And
    Gary Phua Hwee Leng\\
  Information Systems Technology and Design\\
  Singapore University of Technology and Design\\
  \texttt{gphua1@gmail.com} \\
}

\begin{document}
\date{}
\maketitle
\begin{abstract}
Using pretrained word embeddings has been shown to be a very effective way in
improving the performance of natural language processing tasks. In fact almost any
natural language tasks that can be thought of has been improved by these pretrained
embeddings. These tasks range from sentiment analysis, translation, sequence prediction
amongst many others. One of the most successful word embeddings is the
Word2vec CBOW model proposed by Mikolov trained by the negative sampling technique. Mai et al. modifies this objective to train CMOW embeddings that are sensitive to word order. We used a modified version of the negative sampling objective for our context words, modelling the context embeddings as a Taylor series of rate matrices. We show that different modes of the Taylor series produce different types of embeddings. We compare these embeddings to their similar counterparts like CBOW and CMOW and show that they achieve comparable performance. We also introduce a novel left-right context split objective that improves performance for tasks sensitive to word order. Our Word2rate model is grounded in a statistical foundation using rate matrices while being competitive in variety of language tasks.
\end{abstract}

\section{Introduction}

Word embeddings are mathematical representations of words in vector space.  They first took the form of one hot vectors in a local representation, where the size of the vector is equal to the vocabulary size of the language model. All elements take in the value of zero, apart from the index of the vector that corresponds to a word in the vocabulary, which takes the value of one. These representations however, have achieved limited success. It is with the introduction of pretrained word embeddings like Mikolov's Word2vec\cite{mikolov2013efficient} that arguably kickstarted the level of improvements we see in natural language processing today. These word embeddings have a distributed representation where they are encoded with semantic meaning in a geometrical space. This means that words that have similar meaning are close together in vector space. They have also been used to improve the performance of many downstream tasks such as analogy(\cite{mikolov2013efficient}), translation(\cite{zou2013bilingual}, \cite{choi2017context}), sentiment analysis(\cite{yu2017refining},\cite{giatsoglou2017sentiment}), named entity recognition(\cite{demir2014improving}) among many others. While these embeddings have been effective, mathematically they only represent the hidden layer of an autoencoder and do not have much statistical significance. 
For this paper, we will be modelling embeddings as a series of statistical transitions along a Markov chain. We will be using a modified negative sampling objective of Mikolov, that differs in the composition function in forming the context embedding $v_c$. Others such as Mai et al. \cite{mai2019cbow} have attempted to modify the objective before, using matrix multiplication instead of addition as the composition for the context.

Our model uses a Taylor series of rate matrices acting on a initial distribution. for the composition of the context embedding. Depending on the mode of the Taylor series, the context embedding could be composed of addition, product or a mixture of both. We call this novel word embedding Word2rate, and compare the different modes to their existing counterparts. We evaluate these word embeddings on a series of 10 linguistic probing and 11 downstream tasks from Senteval\cite{conneau2018senteval},\cite{conneau2018you}, following Mai. Senteval covers a widerange of tasks, including semantic and syntactic information of a sentence embedding and sentiment classification.

We introduce 3 different modes of the Taylor series from our Word2rate model, namely First Order Series (FOS), First Order Product(FOP) as well as Second Order Series (SOS). We also evaluate 2 Hybrid models formed from these embeddings, namely Word2rate FOS-FOP and Word2rate FOS-SOS, which were formed using the same methods introduced by Mai. We also introduce a novel left-right context split to form context embeddings. We find these embeddings to improve performance on tasks sensitive to word order. We show overall that the different modes of our Word2rate embedding are competitive with their counterparts. At the same time, our model grounds word embedding formation with a statistical foundation by modelling context formation as a series of transitions.

\section{Related Work}
In this paper, we will be training semantic unigram and bigram word embeddings using a model similar to Word2vec. These bigrams will be formed by some composition rules using algebraic operators. Mitchell and Lapata\cite{mitchell2010composition} first gave a mathematical basis to build upon when studying language composition. They first defined a basic composition $p$ as a function two elements u and v:
$$p = f(u; v)$$
These elements $u$ and $v$ can represent words and $p$ can represent a phrase. They then go on to define a more complete composition that will include the syntactic relation R of u and v as well as the existing knowledge used during composition, K:
$$p' = f(u; v;R;K)$$
They studied the performance of various vector composition models on a similarity rating task for phrase pairs in the english language. The aim being able to represent higher level structures of language in a semantic space. Their experiment does not use $K$ as it does not use any outside knowledge during composition. The relationship $R$ that they use is the relationship between the word pairs. They use both semantic and probabilistic models for their word vectors and apply the same algebraic operators to them. These algebraic operators include addition, multiplication, tensor product and dilation among others. For our model, the algebraic operators we use will be mainly addition and matrix multiplication, which involves acting rate matrices on an initial distribution as well combining these operators together. The relationships of the words will be the context-target pair that is the same as Word2vec.

Socher\cite{socher2012semantic} also builds upon the work of Mitchell and Lapata but applies a different approach to language composition. He assigns a matrix $A$ and a vector $a$ to each word. By doing so, 2 words can be described as a cross matrix vector pair:
$$ j = g(W[Ba\; Ab])$$
Where $a$ and $b$ are $n$ dimensional vectors that represent words, with their counterparts $A$ and $B$ being their corresponding matrix.$[Ba\; Ab]$ is the concatenation between the 2 resultant vectors. $g$ is a non-linear function, $W$ is a $2n\times n$ composite matrix that maps the concatenation back to $n$ dimensions, and $j$ is the composition vector of the 2 words. These composition vectors are trained recursively in a tree structure on some supervised sentiment tasks. This tree structure allows the model to capture the syntactic composition of the words.

Mikolov\cite{mikolov2013efficient} produced the seminal paper on word embeddings, training word embeddings on a large unsupervised corpus by splitting words in a set of context-target pairs. There are two different methods to do so, namely skipgram and CBOW. We will focus on CBOW. In CBOW, a collection of context words are summed to form a context embedding $v_c$:

$$v_c = \sum_{i=-m}^{m}W_{c}c_i$$

These context words $w_c$ are within $m$ words of a target word $w_t$. $c_i$ can be thought of as a one-hot vector acting on the list of context embeddings $W_c$. For a corpus with vocabulary size $n$ and embedding dimension $d$, $W_c$ is a $n\times d$ matrix. Forming this context embedding is essentially what differentiates our model from Word2vec. For training does not only involve training individual words, their composition rules or relationships between the words are also incorporated into the embeddings during training. After forming the context embedding, it is then used to predict the target word from a softmax function:
$$p(w_t|w_c) = \frac{v_c^Tv_t}{\sum_{i=1}^{W}\exp(v_i^Tv_t)} $$
Where $v_t$ is the associated target embedding. Unlike the context embedding, it is simply equal to one of the rows of the target embedding matrix $W_t$ instead of a sum. Given a large vocabulary size, computing the softmax directly is computationally expensive. Hence, Mikolov approximates this using a negative sampling objective function:
\[F = \log\sigma(v_t^Tv_c) + \sum_{i=1}^{k}\log\sigma(-v_{nsi}^Tv_c) \tag{2.1}\label{eq:2.1}\]
This negative sampling objective learns to discriminate between good embedding pairs $v_c$, $v_t$ with bad embedding pairs drawn from a unigram distribution, $v_{ns}$ and $v_c$. This will be the objective function that we use for training our embeddings.

Mai et al.\cite{mai2019cbow} on the other hand, trains word embedding matrices on an unsupervised task using a modified version of the context embedding. They essentially replaced the sum operation of the context embedding vector $v_c$ with matrix multiplication. Instead of a vector, they form their context embedding using matrices:
$$M_c = \prod_{i=-m}^{m}W_{ci}$$
Where in this case, $M_c$ is now the context embedding and $W_c$ is now a $n\times d\times d$ context embedding matrix. They show that this is useful as the non-commutating properties of matrix multiplication allow word order to be learnt by the model. This is compared to the usual summing of word vectors, where word order is invariant under summation and will not give the model that information. They also propose using an assymmetric window for the context words by randomly selecting the target word, as well as initialising the embedding matrices with an identity matrix to prevent vanishing gradient problems from multiplication.

Our model shares similar representation to that of Socher, while having a more similar objective function to Mikolov and Mai. We will also be training our model on a modified version of the context embedding. However, by representing our embeddings as statistical operators, we hope to ground our model with a theoretical statistical foundation. In the next section we will introduce the different modes of our Word2rate embeddings as well as our novel left-right context split technique.

\section{Word2rate}
In Word2vec, Mikolov has shown that one can learn word embeddings through negative sampling by maximising the correlation between a correct context-target pair $v_c$ and $v_t$ and penalising the correlation between a wrong context-target pair $v_c$ and $v_ns$. The context word embedding $v_c$ was formed by simply summing the neighboring words embeddings of the target. This summation can be treated as a composition function of the neighboring words. In our model of Word2rate, while using the same negative sampling objective, we employ a different composition function to obtain $v_c$ that represents a Markov chain of transitions from an initial probability distribution. The target embeddings $v_t$ are also unchanged in our model. We decide to use probability distributions as we want to view word embeddings as a distributed representation where each element of the vector can be thought of as some abstract concept state. A word is then uniquely represented by a configuration of these concept states, while a sequence of words can be represented as a transition across multiple states. We first define the initial distribution $p_u$ as a uniform distribution where:
$$p_u \in \mathbb{R}$$
$$(p_u)_i = \frac{1}{d}$$
$$\sum_{i=1}^{d}(p_u)_i = 1$$
Where $d$ is the dimensions of the vector and $(p_u)_i$ is the $i^{\textnormal{th}}$ element of $p_u$. We can then form a word embedding as a transition from the initial distribution. This can be then achieved by applying a stochastic matrix $S_1$ to $p_u$:

$p_1 = S_1p_0$
Where the elements of a stochastic matrix $S$, $s_{ij}$ are conditioned with the following properties:
\[s_{ij}\leq 0\tag{3.1}\label{eq:3.1}\]
\[\sum_{i=1}s_{ij}=1\tag{3.2}\label{eq:3.2}\]

Equation \ref{eq:3.1} is conditioned due to the elements of the stochastic matrices being probabilities and hence have to be non-negative. Equation \ref{eq:3.2} is conditioned for each column to sum to 1 to conserve probability as a left stochastic matrix which acts on the left of the probability distribution. Having introduced the properties of stochastic matrices, we can then define our word embedding as the stochastic matrix of a particular word acting on a distribution. To form a word embedding like 'cat', we can apply the corresponding stochastic matrix $S_{cat}$ to the initial distribution $p_u$
$$v_{cat} = S_{cat}p_u$$
We can also defined a sequence of words as series of stochastic matrices acting on an initial distribution. We will demonstrate this with an example of a sentence 'We love cats'. We can let 'cats' be the target word while the phrase 'We love' is the context. We can then represent the context embedding as:
\[v_c = S_{love}S_{We}p_u \tag{3.3}\label{eq:3.3}\]

In order to gain a richer representation of the composition function, we can approximate these stochastic matrices using a Taylor series. Firstly, a stochastic matrix $S$ can be expressed as an exponential function of a rate matrix $Q$:
$$S = \exp(\epsilon Q)$$
Where $\epsilon \ll 1$. Since $Q$ is a matrix, expressing it as an exponential is non-trivial. We can typically do so by approximating the exponential function as a Taylor expansion:
\[ S = I + \epsilon Q + \frac{\epsilon^2 Q^2}{2!} + ... \tag{3.4}\label{eq:3.4}\]
We can then apply Equation \ref{eq:3.4} above to Equation \ref{eq:3.3}:
\[\small v_c = (I + \epsilon Q_{love} + \frac{\epsilon^2 Q_{love}^2}{2!}+...)(I + \epsilon Q_{We} + \frac{\epsilon^2 Q_{We}^2}{2!}+...)p_u \tag{3.5$\ast$}\label{eq:3.5}\]

Equation \ref{eq:3.5} is the crux of our Word2rate model. We can see a variety of ways that we can choose to approximate this infinite expansion. Depending on how we choose to approximate, we will get different composition rules which will provide different qualities for embeddings trained under these rules.

\subsection{Word2rate First Order Series(FOS)}

\[v^{FOS}_c = (I + \epsilon Q_{love} + \epsilon Q_{We})p_u \tag{3.6}\label{eq:3.6}\] 

We first approximate the Taylor series using only up to the first order of $\epsilon$. We illustrate this in equation \ref{eq:3.6} by continuing our 'We love' example earlier on. We can see from the equation that our FOS embeddings are formed using only addition between rate matrices. This form of additive composition is the same as that used by Mikolov for Word2vec. Hence we can say that

\subsection{Word2rate First Order Product(FOP)}
\[v^{FOP}_c = (I + \epsilon Q_{love})(I + \epsilon Q_{We})p_u \tag{3.7}\label{eq:3.7}\] 
Instead of approximating the entire Taylor series up to the first order, we can approximate the individual product terms to the first order instead. This would form our Word2rate FOP context embedding in this manner. We can first observe that this form of approximation would allow the context embedding to contain many more terms in the expansion than our FOS embeddings implicitly. For our FOS embeddings or CBOW, the number of terms of the embedding scale linearly with the number of context words. For our 2 word example, we have 2 terms in the expansion comprising of the rate matrix $Q_{love}$, $Q_{We}$ as well as the identity matrix. For our FOP embeddings, we will have 2 term, the product contains 4 terms, including the bilinear term $\epsilon Q_{love}Q_{We}$. This will allow us to train a relationship between the words `We' and `love' that incorporates the syntactic properties of the sequence.  This respects the order that `love' comes after `We' in this case, since our rate matrices like CMOW are sensitive to word order. However, unlike CBOW our CMOW, our expansion implicitly contains $2^{2m}$ terms with $m$ being the window size after expansion. This is due to the addition of the identity matrix, which allows for cross product between words. While CMOW introduces the identity matrix as an engineering technique to avoid the vanishing gradient problem, our identity matrix is a mathematical feature arising from statistical theory, and it is added independently of the trained rate matrices. We will later show in our experiments the role of $\epsilon$ in avoided vanishing gradients instead. Furthermore, unlike CMOW, our matrix embeddings do not unroll to form word embeddings. Instead, they act on an initial uniform distribution and transit that initial state as it progresses through the sequence of context to form our final context embedding.

\subsection{Left-Right Context Split}
CBOW and CMOW use a combination of a sequence of words on both the left and right side of the target word to form their context embeddings. For example, for a sentence like `The quick fox jumps', if the target word is `fox', the the context embedding can be formed as:
$$v_c = f(w_{the}, w_{quick}, w_{jumps})$$
Where $f$ could be the sum operation of CBOW, product operation of CMOW or one of our Word2rate series operations. In all cases, the context embedding is formed as `The quick jumps'. Recall earlier that we mentioned that the context embedding learns the relationships between words in a sequence as well. This is why sentence embeddings for downstream tasks are formed by adding CBOW word embeddings, while being formed by multiplying CMOW embeddings instead. However, a problem arises as `The quick jumps' is not syntactically correct in the english language. Hence, this will cause inaccuracies especially when using reprensentations like matrices which are sensitive to word order. To solve this issue, we propose a novel left-right context split that separates the words appearing on the left and right of the target word as 2 separate embeddings. For this example, the left and right context embeddings would take the form $v_{cl}=f('w_{the}, w_{quick})$ and $v_{cr} = f(w_{jumps})$ respectively. After separation, our negative sampling objective will now be a modified version of equation \ref{eq:2.1}:

\begin{align*}
F &= \log\sigma(v_t^Tv_{cl}) + \sum_{i=1}^{k}\log\sigma(-v_{nsi}^Tv_{cl}) \tag{3.8} \label{eq:3.8}
+ \log\sigma(v_t^Tv_{cr}) + \sum_{i=1}^{k}\log\sigma(-v_{nsi}^Tv_{cr})
\end{align*}
We will be using this new objective function in \ref{eq:3.8} to train our left-right context embeddings. While we change the training objective, we still use draw the word embeddings from the same context embedding matrix $W_c$.
\subsection{Hybrid Embeddings}
Hybrid embeddings was a technique first proposed by Mai. These embeddings involve training both their CBOW and CMOW embeddings concatenated together as a context embedding. In their paper they argue that training the embeddings together helps to remove redundancies which would occur if they were trained separately and then concatenated afterward. The representation of their hybrid context embedding $v_{ch}$is written as:
\[v_{ch} = [v_{cmow}; v_{cbow}]\tag{3.9}\label{eq:3.9}\]
Where $v_{cmow}$ and $v_{cbow}$ are the context embeddings formed by their respective operations. We will be using the same form of hybrid embeddings as equation $3.9$, but replacing $v_{cbow}$ and $v_{cmow}$ with our own Word2rate embeddings.
\subsection{Word2rate Second Order Series}

\begin{align*}
\small v^{SOS}_c &= (I + \epsilon Q_{love} + \epsilon Q_{We} + \epsilon^2 Q_{love}Q_{We} 
+ \frac{\epsilon^2 Q_{love}^2}{2!} +  \frac{\epsilon^2 Q_{We}^2}{2!} )p_u \tag{3.10}\label{eq:3.10}
\end{align*}
We can take all terms up to the second order from the Taylor series expansion in equation \ref{eq:3.5} to form our Word2rate Second Order Series(SOS) represented in equation \ref{eq:3.10}. From the equation we can see that compared to our FOS embeddings, there are exists the bilinear term $\epsilon^2 Q_{love}Q_{We}$. Also unlike our FOP embeddings this term is now explicit in nature. As we have discussed earlier, having these bilinear interactions in matrix form allows us to learn relationships between words. At the same time, our SOS context embedding also contains the original terms from our FOS embeddings. Hence we can say that it learns relationships between word embeddings using properties from both CBOW and CMOW. 
\begin{table*}[!h]
	\begin{center}
		\begin{adjustbox}{width=1\textwidth}
			
			\begin{tabular}{c|c|c|c|c|c|c|c|c|c|c|c}
				\def\arraystretch{1.2}
				{} & $\epsilon$ & {Length} & {WC} & {Depth} &{TopConst} & {Bshift} & {Tense} & {SubjNum} &{ObjNum} & {SOMO} &{Coordinv} \\
				\hline
				CBOW 25             & -      & 24.8 & 23.0 & 21.8 & 20.5 & 50.1 & 71.8 & 74.0 & 75.1 & 50.0 & 51.5 \\
				CMOW 25             & -      & 47.5 & 13.7 & 22.6 & 20.3 & 50.7 & 67.4 & 67.2 & 66.6 & 50.9 & 51.8 \\
				Word2rate FOS 25    & 0.01   & 59.0 & 18.5 & 24.2 & 23.8 & 50.5 & 75.2 & 74.7 & 76.0 & 49.6 & 51.8 \\
				Word2rate FOP       & 0.001  & 29.4 & 13.5 & 22.2 & 18.4 & 53.6 & 68.7 & 66.8 & 68.3 & 50.6 & 53.3 \\
				Word2rate FOP 25 lr & 0.001  & 24.8 & 10.1 & 20.8 & 18.2 & 58.2 & 64.0 & 62.5 & 63.3 & 50.9 & 54.8 \\
				Word2rate SOS 25    & 0.001  & 66.7 & 8.9  & 23.9 & 18.7 & 50.7 & 68.9 & 66.2 & 68.1 & 49.2 & 55.1 \\
				CBOW 50             & -      & 25.3 & 42.3 & 23.1 & 24.0 & 49.8 & 76.3 & 75.1 & 78.5 & 49.9 & 52.1 \\
				Hybrid FOS-FOP 50   & 0.0001 & 44.1 & 29.8 & 23.9 & 25.1 & 50.9 & 76.0 & 72.9 & 75.5 & 50.4 & 55.1

			\end{tabular}
		\end{adjustbox}

		\caption{Results of linguistic probing tasks from Senteval}
		\label{table:lp}
	\end{center}
\end{table*}

\begin{table*}[h!]
	\begin{center}
		\begin{adjustbox}{width=1\textwidth}
			
			\begin{tabular}{c|c|c|c|c|c|c|c|c|c|c|c|c}
				\def\arraystretch{1.2}
				
				& $\epsilon$ & MR   & CR   & MPQA & SUBJ & SST2 & SST5 & TREC & MRPC & SICK-E & SICK-R & STS-b \\
				\hline
				CBOW 25             & -      & 64.3 & 69.7 & 79.4 & 84.9 & 64.7 & 33.2 & 57.0 & 70.4 & 65.8 & 62.9 & 48.6 \\
				CMOW 25             & -      & 59.0 & 64.3 & 77.0 & 74.2 & 59.5 & 30.2 & 56.4 & 67.1 & 62.7 & 54.8 & 24.6 \\
				Word2rate FOS 25    & 0.01   & 65.1 & 70.3 & 77.4 & 84.2 & 67.1 & 33.9 & 57.2 & 71.5 & 69.7 & 62.3 & 45.2 \\
				Word2rate FOP  25   & 0.001  & 60.0 & 63.8 & 69.2 & 77.7 & 60.6 & 30.4 & 54.0 & 66.5 & 58.0 & 52.3 & 45.5 \\
				Word2rate FOP 25 lr & 0.001  & 57.2 & 63.8 & 70.9 & 73.6 & 55.1 & 26.2 & 48.8 & 66.5 & 58.3 & 50.5 & 38.3 \\
				Word2rate SOS 25    & 0.001  & 61.2 & 65.6 & 68.8 & 78.9 & 61.9 & 32.0 & 50.2 & 68.6 & 57.0 & 50.2 & 36.3 \\
				CBOW 50             & -      & 68.1 & 72.4 & 82.6 & 85.9 & 70.5 & 35.6 & 59.2 & 71.4 & 70.6 & 65.1 & 53.4 \\
				Hybrid FOS-FOP 50   & 0.0001 & 63.9 & 71.0 & 79.3 & 83.3 & 65.7 & 34.1 & 59.4 & 70.5 & 72.7 & 63.9 & 53.1

			\end{tabular}
		\end{adjustbox}
		
		\caption{Results of downstream tasks from Senteval}
		\label{table:ds}
	\end{center}
\end{table*}
\section{Experiments}
For our experiments, we will be training different modes of the Word2rate Taylor series that we just described while comparing them to CBOW and CMOW. We will also conduct various ablation studies on these embeddings. These embeddings will be trained on 50 percent of the billion words dataset with vocabulary consisting of words that occur at least 100 times and sentence length between 10-20 words. This is equal to a corpus with 5982756 sentences with vocabulary size 57025. We will be using the Adam(\cite{kingma2014adam}) optimiser with learning rate 0.001, batch size 1000 and number of epochs equal to 10. We will train embeddings of dimension 25 for CBOW, CMOW, FOS, FOP and SOS while training 50 dimensional embeddings for our hybrid models. The window size for the context words will be set to 4 while the size of the negative samples per context-target pair will be set to 5. While the elements of the rate matrices $Q$ will be altered during gradient descent, we enforce the rate matrix conditions manually after each training step to make sure that they stay continue to fulfil the conditions for a rate matrix. 
We will evaluate these embeddings on a number of tasks, from Senteval. The Senteval framework that we will be using consists of 10 linguistic probing tasks as well as 11 supervised downstream tasks. The 10 linguistic probing tasks can be divided into 3 categories, namely surface information, syntactic information as well as semantic information of a sentence. The 11 downstream tasks mostly involve some kind of classification tasks, with datasets ranging from entity recognition(TREC) to sentiment classification(CR, MR etc.). As we have mentioned earlier, we will form the sentence embeddings the same way we train our context embeddings. For tasks that are sensitive to word order and require product, we use the assymmetric context window proposed by Mai, and also use their initialisation for evaluating the CMOW embeddings. While CMOW randomly selects 30 words as targets from each sentence, since we are using a relatively smaller dataset we randomly select $n$ words for a sentence of length $n$ as the target instead. We present our results for the linguistic probing and downstream tasks in Tables \ref{table:lp} and \ref{table:ds} respectively.

\subsection{Word2rate FOS}
We first start by comparing our Word2rate FOS embeddings with CBOW since they both use the same add operation in composing the context embedding. For the linguistic probing tasks, we can see that our FOS embeddings outperforms CBOW in eight out of 10 of the tasks, losing slightly in SOMO and noticeably in WC by 4.5 percent. Where WC is a word memorisation tasks. FOS performs significantly better in Length by about 30 percent, which involves predicting the number of words a sentence possessed based on its sentence vector.
For the downstream tasks, FOS performs better in 7 out of the 11 tasks, while being within 1 percent accuracy for the SUBJ and SICK-R tasks.
One likely explanation for the improve in performance could be due to the richer representation of our FOS embedding as each $n$ dimensional embedding is formed by a $n \times n$ rate matrix multiplied by the uniform distribution $q_u$. Training this higher dimensional representation and then compressing it to a lower dimensional embedding could explain the better performance for the Length task.

On the flipside, one possible limitation of our FOS embeddings is in the relatively higher training complexity compared to the regular CBOW embedding. To produce a $n$ dimensional word embedding $q_w$ we also have to train its corresponding rate matrix $Q_w$ which is $n \times n$ dimensions. Even though training time is longer, the trained word2rate embeddings have the same dimensionality as that of word2vec. As such there will not be an increase in complexity when using them on downstream tasks. It is also important to note that our FOS embeddings and CBOW use the same number of trainable parameters for downstream tasks. The only difference is in the way we form the sentence vector.

We have shown experimentally that our Word2rate FOS embeddings is competitive and even outperforms CBOW in most linguistic probing and downstream tasks, likely due to our compressed representation of rate matrix-vector pairs.
\subsection{Word2rate FOP}
\begin{figure}[!htbp]
	\centering
	\includegraphics[width=1\linewidth]{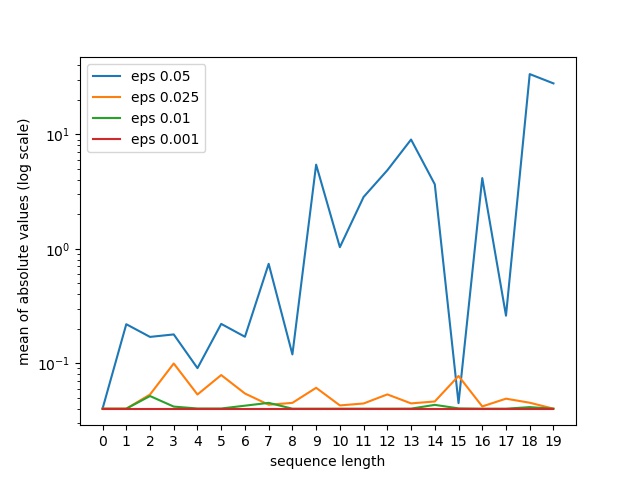}
	\caption{Mean of absolute values of matrix elements}
	\label{fig:meanplot}
\end{figure}
In the previous section, the value of the hyperparameter $\epsilon$ was kept constant at 0.01. Since addition is linear, varying the $\epsilon$ does not make a difference in performance which we show in the Appendix. However, $\epsilon$ will have an impact in performance for FOP due to the nature of the matrix product. In the case of Mai, there arises vanishing gradient problem due to elements of the matrix initialising around a mean of 0. This is not applicable to us as we add the identity to the rate matrix for each product term. However, we still see an impact in $\epsilon$ as we can see in Figure \ref{fig:meanplot}.  This Figure is a plot of the mean absolute value of a rate matrix product after being multiplied by number of times equal to sequence length. This is the same experiment conducted by Mai, but using our FOP composition. From the experiment we can see a clear trend that the magnitude of fluctuations decreases with smaller values of $\epsilon$, being relatively stable at $\epsilon=0.001$. Stable values is good for training as it provides more stable gradient updates. Hence we will be using $\epsilon$ values of 0.001 or lower for tasks involving matrix product. 
\subsubsection{Comparison with CMOW}
We can next refer to Tables \ref{table:lp} and \ref{table:ds} and compare our FOP embeddings with CMOW as both involve matrix product. For linguistic probing tasks, FOP outperforms CMOW in 4 out of the 10 tasks, while bieng within 1 percent accuracy in 4 other tasks, WC, Depth, SUBJ and SOMO. However, it performs noticeably worse in Length. This is in contrast to our FOS embeddings which outperforms both CBOW and CMOW in the Length tasks. This seemingly contradicts our hypothesis that a compressed representation would allow for better performing embeddings. However, one possible reason for the drop in performance could be that our FOP embeddings implicitly contains a multitude of terms, up to $2^{2m}$ with $m$ being the window size. This large number of terms could possibly be interpreted as noise during training which could explain the dip in performance.
For the downstream tasks, FOP in 5 out of the 11 tasks while being within 1 percent accuracy for 2 of the tasks. The largest drop in performance is in MPQA with CMOW outperforming FOP by close to 9 percent, where MPQA is an opinion polarity classification task. On the other hand, FOP outperforms CMOW by more than 20 percent on STS-b, a semantic similarity tasks. While we cannot pinpoint the exact reason for the variance in these 2 tasks, it could be due to the nature of the dataset, or the noise and compressed representation factors we mentioned earlier. While there are some differences in performance, most of the tasks are relatively even with CMOW, and we have shown that our rate matrix-vector combination is comparable to the matrix unrolling representation of CMOW.
\subsubsection{Comparison with CBOW}
\raggedbottom 
In linguistic probing, FOP outperforms CBOW in 5 out of the 10 tasks. Notably, FOP outperforms CBOW in Bshift and Coordinv by 2 and 3 percent respectively. These 2 tasks are sensitive to word order, more specifically, Bshift involves detecting when there is a swapping of words while Coordinv involves swapping of phrases. Usual additive composition employed by CBOW is unable to detect such changes as addition is invariant to order. This is one of the benefits of matrix multiplication, which is sensitive to order and non-commutable. These properties are also present in CMOW which also outperforms CBOW in these tasks.
FOP fairs much more poorly in downstream tasks, losing to CBOW across all tasks. However, this is not surprising and a similar behavior can also be observed with CMOW embeddings in Table \ref{table:ds}. Furthermore, this is behavior is also somewhat observed in the original CMOW paper by Mai, where his embeddings lose to CBOW on 9 out of the 11 downstream tasks. This seems to be a property of using multiplication to compose context embeddings, where addition seems to have an advantage in downstream tasks. This could be due to the geometric properties of addition, where the summed context embedding is similar to its constituent words. This is opposed to matrix product which is more of a transformation of the embedding from one space to another. To alleviate this issue, Mai proposes using a hybrid training method of both CBOW and CMOW. We will train a Hybrid FOS-FOP embedding with the same method and evaluate its performance later on. For now, we will introduce our novel left-right context split which we used to improve performance for our FOP embeddings in some tasks.

\subsubsection{Left-Right Context Split}
We introduced our novel left-right context (lrcs) split previously in Section 3.3. We can see the performance under `Word2rate FOP 25 lr' in Tables \ref{table:lp} and \ref{table:ds}. We will be comparing its performance with the usual FOP embeddings in the previous section. For the linguistic probing tasks, we can see that our lrcs embeddings outperforms FOP in 3 of the 11 tasks, namely Bshift, SOMO and Coordinv. Bshift and Coordinv notably are sensitive to word order, and our lrcs embeddings outperforms FOP by 4.6 and 1.5 percent on these two tasks respectively. This shows that adding the left-right context split does improve performance of word order due to preserving syntactically correct sequences. However, our lrcs embeddings does overall perform worse in both linguistic probing and especially downstream tasks, performing only slightly better or tieing in 4 out of the 11 tasks.

We think there could be 2 possible reasons for the drop in performance compared to FOP. Firstly, a left-right split means the embeddings learn a shorter sequence length in some cases. If the target word is at the edge of the window, there is no difference. However, if there target word is in the middle then the context embedding formed without the split would be longer compared to a left-right split. Training a longer context embedding instead of two shorter ones may result in better performance in downstream tasks. 

Secondly, while forming a sentence embedding for downstream tasks, there are out of vocabulary words present depending on the tasks and respective dataset. These out of vocabulary words are typically padded with an identity matrix. However, a sentence formed with out of vocabulary words could also be syntactically inaccurate. Consider the previous example of `the quick fox jumps'. We have illustrated earlier that without the left-right context split the model would learn the context embedding `the quick brown jumps'. However, if the word `fox' were to be replaced by an out of vocabulary word like `florpus' for the downstream task, then the resultant sentence vector would also be 'the quick jumps'. This negates some of the potential impact of performing the left-right split.

\subsection{Hybrid FOS-FOP}
We evaluated our Hybrid FOS-FOP against CBOW, CMOW and FOP in 50 dimensions to see if they would overcome the limitations of our matrix product model for FOP, especially for the downstream tasks. Firstly we can see for our linguistic probing tasks that our Hybrid embeddings outperform CBOW in 5 out of 10 tasks. As usual our Hybrid model is better than CBOW in Bshift and Coordinv. However, it is worth noting that the accuracy for Bshift is lower than our FOP embeddings by 2.7 percent. Training a mixture of FOS and FOP could possibly negate part of the benefits of either model.

We also compared our Hybrid embedding with CMOW and FOP at 50 dimensions. We can see that having a Hybrid model does not degrate its performance in linguistic probing task. From Table \ref{table:lp} we can see that our Hybrid model outperforms FOP in 8 out of 10 tasks and CMOW in 7 out of the 10 tasks. Compared to CMOW, the tasks Hybrid loses on differ in accuracy by close to 1 percent. For CMOW, Hybrid fairs worse notably in Bshift losing by 2.7 percent. This is likely due to the mixture of linear and product operations reducing the signal of word sensitivity from the matrix products.

For the downstream tasks, we can see that our Hybrid embeddings doing better than FOP and CMOW across all tasks,  while beating CBOW in 2 tasks and being within 1 percent accuracy in 2 other tasks.  This shows that training a Hybrid model does help strengthen the performance in tasks where each individual embedding is lacking. However, it could also deter its performance in tasks where they are strong as well. The expanded version of FOP contains all terms from FOS implicitly. This could result in a repetition during training. To investigate this we trained a Hybrid FOS-FOP minus embedding where we subtracted the FOS terms from FOP. We will discuss this further in the Appendix.

\subsection{Word2rate Second Order Series}
Our Word2rate SOS embeddings contain both addition and product in the formation of the context embedding while containing bilinear interactions between rate matrices. Even though there are implicit bilinear interactions in FOP, there are also many additional terms higher than the second order. Limiting ourselves to only second order terms allows us to investigate the impact of the high order terms as well. Since SOS contains both addition and product, we will be comparing it with the respective CBOW and CMOW embeddings.

\subsubsection{Comparison with CBOW}
For the linguistic probing tasks, we can see that SOS outperforms CBOW in 4 out of 10 tasks. SOS still retains the properties of our usual word2rate embeddings, outperforming CBOW in Length by a large margin of more than 40 percent. It also fairs better in both Bshift and Coordinv. This is likely due to the bilinear interactions of the second order terms of the rate matrix. However, similar to our hybrid embeddings it fairs better for Coordinv but worse for Bshift compared to our FOP embeddings. The drop in performance for Bshift is likely due to the presence of linear terms which are invariant to word order. However, we are unsure of why these embeddings do perform better for Coordinv. One possible reason could be that Coordinv measure phrase rather than word order and having addition as part of phrase formation has a positive impact. For the downstream tasks we observe that CBOW outperforms SOS in all tasks. The bilinear product terms are now the ones causing a dip as we observe addition only for CBOW to perform better overall for the downstream tasks. 

\subsubsection{Comparison with CMOW}
Our SOS embeddings are roughly equal to CMOW for the linguistic probing tasks, with SOS outperforming SOS in 5 out of the 10 tasks while drawing in 1 of them with most accuracies being fairly close. Our SOS embeddings continue to have good performance in Length, while losing out to both CBOW and CMOW in WC which is a memorisation tasks. This drop in performance in WC is not due to using bilinear terms as our FOP embeddings have roughly the same result as CMOW. It could be likely that having a mixture of linear and bilinear terms is not that good for memorisation.

We see a marked increase in performance of SOS against CMOW for the downstream tasks. Our previous FOP embeddings are only better in 5 out of 11 tasks compared to CMOW. However SOS outperforms CMOW in 7 out of the 11 tasks. This increase is likely due to the presence of the addition operation. We have also previously discussed that the drop in performance of Length for our FOP embeddings is due to the multitude of high order terms. We can see that this may be the case as our SOS embeddings with only bilinear interactions actually is the best performing in the Length tasks, beating both FOS and FOP. This is interesting as we have earlier discussed along with our Hybrid embedding that addition and product together helps each others weakness while likely decreasing their strength in certain tasks. Length is an example where having both addition and product increases overall performance.

\section{Conclusion}

In conclusion, we have show that using our novel Word2rate FOS, FOP, SOS as well as Hybrid embeddings are effective feature representations and that are competitive with existing embeddings on a number of tasks. In other words, it is feasible to model word embeddings as a series of transitions from an initial distribution. We have shown how these word embeddings can be formed by taking the first order, second order series or product approximation of our Taylor expansion of rate matrices. The FOS embeddings have shown comparable performance to the existing word2vec model on a number of linguistic probing as well asdownstream tasks. 
We have also shown that FOP embeddings, while having some limitations, has shown promising properties that make it a better candidate than the word2vec addiitive composition model when forming phrases. Their sensitivity to word order is also useful in linguistic probing tasks such as Bshift and Coordinv where they outperform CBOW. 

We also introduce a novel left-right context split of the context embedding that further improves the performance of such tasks. While our model does share some similarities with CMOW, our model uses rate matrices rather than random matrix which are constrained in their diagonal elements. This allows us to ground our embeddings in a statistical foundation. Our embeddings are also formed from a combination of rate matrix-vector pair instead of the unrolling operation employed by CMOW.

At the same time we have shown that combining our Word2rate FOS and FOP embeddings as a hybrid model allows us to show comparable performance to CBOW embeddings with the same dimensions on the downstream tasks. On the other hand, the second order series Taylor expansion seems to be a mixture the first order series and the first order product, sharing both sum and product matrix operations. While it has shown to have outperform both CBOW and CMOW in some tasks, it does not perform as well as training the sum and product terms individually like in our FOS and FOP embeddings.

In essence, we have proposed a novel way of representing a language model and in a deeper sense a model of how we form concepts in our mind. This is modelled as a series of statistical transitions from an initial distribution can be interpreted as the evolution of our thoughts.

\bibliography{reference}

\begin{thebibliography}{10}

\bibitem{mikolov2013efficient}
Tomas Mikolov, Kai Chen, Greg Corrado, and Jeffrey Dean.
\newblock Efficient estimation of word representations in vector space.
\newblock {\em arXiv preprint arXiv:1301.3781}, 2013.

\bibitem{zou2013bilingual}
Will~Y Zou, Richard Socher, Daniel Cer, and Christopher~D Manning.
\newblock Bilingual word embeddings for phrase-based machine translation.
\newblock In {\em Proceedings of the 2013 Conference on Empirical Methods in
  Natural Language Processing}, pages 1393--1398, 2013.

\bibitem{choi2017context}
Heeyoul Choi, Kyunghyun Cho, and Yoshua Bengio.
\newblock Context-dependent word representation for neural machine translation.
\newblock {\em Computer Speech \& Language}, 45:149--160, 2017.

\bibitem{yu2017refining}
Liang-Chih Yu, Jin Wang, K~Robert Lai, and Xuejie Zhang.
\newblock Refining word embeddings for sentiment analysis.
\newblock In {\em Proceedings of the 2017 conference on empirical methods in
  natural language processing}, pages 534--539, 2017.

\bibitem{giatsoglou2017sentiment}
Maria Giatsoglou, Manolis~G Vozalis, Konstantinos Diamantaras, Athena Vakali,
  George Sarigiannidis, and Konstantinos~Ch Chatzisavvas.
\newblock Sentiment analysis leveraging emotions and word embeddings.
\newblock {\em Expert Systems with Applications}, 69:214--224, 2017.

\bibitem{demir2014improving}
Hakan Demir and Arzucan {\"O}zg{\"u}r.
\newblock Improving named entity recognition for morphologically rich languages
  using word embeddings.
\newblock In {\em 2014 13th International Conference on Machine Learning and
  Applications}, pages 117--122. IEEE, 2014.

\bibitem{mai2019cbow}
Florian Mai, Lukas Galke, and Ansgar Scherp.
\newblock Cbow is not all you need: Combining cbow with the compositional
  matrix space model.
\newblock {\em arXiv preprint arXiv:1902.06423}, 2019.

\bibitem{conneau2018senteval}
Alexis Conneau and Douwe Kiela.
\newblock Senteval: An evaluation toolkit for universal sentence
  representations.
\newblock {\em arXiv preprint arXiv:1803.05449}, 2018.

\bibitem{conneau2018you}
Alexis Conneau, Germ{\'a}n Kruszewski, Guillaume Lample, Lo{\"\i}c Barrault,
  and Marco Baroni.
\newblock What you can cram into a single vector: Probing sentence embeddings
  for linguistic properties.
\newblock {\em arXiv preprint arXiv:1805.01070}, 2018.

\bibitem{mitchell2010composition}
Jeff Mitchell and Mirella Lapata.
\newblock Composition in distributional models of semantics.
\newblock {\em Cognitive science}, 34(8):1388--1429, 2010.

\bibitem{socher2012semantic}
Richard Socher, Brody Huval, Christopher~D Manning, and Andrew~Y Ng.
\newblock Semantic compositionality through recursive matrix-vector spaces.
\newblock In {\em Proceedings of the 2012 joint conference on empirical methods
  in natural language processing and computational natural language learning},
  pages 1201--1211. Association for Computational Linguistics, 2012.

\bibitem{kingma2014adam}
Diederik~P Kingma and Jimmy Ba.
\newblock Adam: A method for stochastic optimization.
\newblock {\em arXiv preprint arXiv:1412.6980}, 2014.

\end{thebibliography}
\bibliographystyle{unsrt}

\end{document}